# A Gabor block based Kernel Discriminative Common Vector (KDCV) approach using cosine kernels for Human Face Recognition.

Arindam Kar[1+], Debotosh Bhattacharjee[2], Dipak Kumar Basu[2*], Mita Nasipuri[2], Mahantapas Kundu[2]

[1] *Indian Statistical Institute, Kolkata-700108, India*
*+ Corresponding author*
[2] *Department of Computer Science and Engineering, Jadavpur University, Kolkata- 700032, India*
*\* Former Professor & AICTE Emeritus Fellow*
*Email: {kgparindamkar@gmail.com, debotosh@indiatimes.com, dipakkbasu@gmail.com, m.nasipuri@cse.jdvu.ac.in, mkundu@cse.jdvu.ac.in}*

**Abstract**

In this paper a nonlinear Gabor Wavelet Transform (GWT) discriminant feature extraction approach for enhanced face recognition is proposed. Firstly, the low-energized blocks from Gabor wavelet transformed images are extracted. Secondly, the nonlinear discriminating features are analyzed and extracted from the selected low-energized blocks by the generalized Kernel Discriminative Common Vector (KDCV) method. The KDCV method is extended to include cosine kernel function in the discriminating method. The KDCV with the cosine kernels is then applied on the extracted low energized discriminating feature vectors to obtain the real component of a complex quantity for face recognition. In order to derive positive kernel discriminative vectors; we apply only those kernel discriminative eigenvectors that are associated with non-zero eigenvalues. The feasibility of the low energized Gabor block based generalized KDCV method with cosine kernel function models has been successfully tested for image classification using the $L_1$, $L_2$ distance measures; and the cosine similarity measure on both frontal and pose-angled face recognition. Experimental results on the FRAV2D and the FERET database demonstrate the effectiveness of this new approach.

## 1. Introduction

Face authentication has gained considerable attention in the near past through the increasing need for access verification systems using several modalities like voice, face image, fingerprints, pin codes etc. Such systems are used for the verification of a user's identity on the Internet, when using automated banking system, or when entering into a secured building, and so on. The Gabor wavelet transformation (GWT) models well the receptive field profiles of the cortical simple cells, and also has the properties of multi-scale and multidirectional filtering. These properties are in accordance with the characteristics of human vision [1, 2, 3]. Further, the discriminant analysis is an effective image feature extraction and recognition technique as they allow the extraction of discriminative features, reduce dimensionality, and consume less computing time [4, 5]. In our previous work [6], we combined the GWT and Bayesian principal component analysis (PCA) techniques and presented a GWT-Bayesian PCA face recognition method which outperforms some conventional linear discriminating methods. As an extension of linear discriminant technique, the kernel based nonlinear discriminant analysis technique has now been widely applied to the field of pattern recognition. Baudat and Anouar [7] developed a commonly used generalized discriminant analysis (GDA) method for nonlinear discrimination. Jing et al. [8] put forward a Kernel Discriminative Common Vectors (KDCV) method. . In this paper we develop block- based GWT KDCV and propose a block-based low energized nonlinear GWT discriminant feature extraction for enhanced face recognition. As the high energized blocks of GWT image generally have larger nonlinear discriminability values. Then the non-linear discriminant features are extracted from the selected low-energized block of GWT image by presenting a new generalized KDCV method is then extended to include cosine kernel model which extracts the nonlinear discriminating features from the selected blocks to get the best recognition result. These features are finally used for classification using three different classifiers. The experimental results demonstrate the effectiveness of this new approach.

In this paper a novel method is proposed based on selecting low energized blocks of Gabor wavelet responses as feature points, which contain discriminate facial feature information, instead of using predefined graph nodes as in elastic graph matching (EGM) [9], which reduces representative capability of Gabor wavelets. This corresponds to enhancement of edges for eyes, mouth, nose, which are supposed to be the most important points of a face; hence the algorithm allows these facial features to keep overall face information along with local characteristics.

The remainder of this paper is organized as follows: Section 2 describes the derivation of low-energized blocks from the GWT images. Section 3 details the generalized KDCV method with cosine kernel function for enhanced face recognition. Section 4 shows the performance of the proposed method on the face recognition by applying it on the datasets from the FERET [10], and FRAV2D [11] face databases, and by comparing it with some of the previous KDCV method and we conclude our paper in Section 5.

## 2. 2D Gabor Wavelets

Gabor wavelets are used in image analysis because of their biological relevance and computational properties [12, 13]. The Gabor transform is suitable for analyzing gradually changing data such as the face, iris and eyebrow images. The Gabor filter used here has the following general form:

$$\varphi_{\mu,\nu}(z) = \frac{\|k_{\mu,\nu}\|^2}{\sigma^2} e^{-\|k_{\mu,\nu}\|^2 \|z\|^2 / 2\sigma^2} \left[ e^{ik_{\mu,\nu}z} - e^{-(\sigma^2/2)} \right] \quad (1)$$

where $\mu$ and $\nu$ defines the orientation and scale of Gabor kernels respectively, $z = (x, y)$, is the variable in spatial domain, $\|.\|$ denotes the norm operator, and $k_{\mu,\nu}$ is the frequency vector which determines the scale and orientation of Gabor kernels, $k_{\mu,\nu} = k_\nu e^{i\phi_\mu}$ where $k_\nu = k_{max}/f^\nu$ and $k_{max} = \pi/2$, $\phi_\mu = \pi\mu/8$, $\mu = 0,...7$, where $f$ is the spacing factor. Here Gabor wavelets at five different scales, $\nu \in \{0,..., 4\}$ and eight orientations $\mu \in \{0,..., 7\}$ are chosen. The term $e^{-(\sigma^2/2)}$ is subtracted from equation (1) in order to make the kernel DC-free, thus becoming insensitive to illumination. The magnitude of the convolution outputs are indicated as $o_{\mu,\nu}(z)$. The kernels exhibit strong characteristics of spatial locality and orientation selectivity, making them a suitable choice for image feature extraction when one's goal is to derive local and discriminating features for (face) classification.

*2.1. Gabor based feature Representation*

The Gabor wavelet representation of an image is the convolution of the image with a family of Gabor kernels as defined in (1). Let $I(x, y)$ be the gray level distribution of an image, the convolution output of image $I$ and a Gabor kernel $\varphi_{\mu,\nu}$ is defined as:

$$O_{\mu,\nu}(z) = I(z) * \varphi_{\eta,\nu}(z), \quad (2)$$

where z = (x, y), and * denotes the convolution operator. Applying the convolution theorem, convolution outputs are derived from (2) via the Fast Fourier Transform (FFT):

$$\Im\{O_{\mu,\nu}(z)\} = \Im\{I(z)\}\Im\{\varphi_{\mu,\nu}(z)\}, \quad (3)$$

and, $O_{\mu,\nu}(z) = \Im^{-1}\{\Im\{I(z)\}\Im\{\varphi_{\mu,\nu}(z)\}\}, \quad (4)$

Here $\Im$, $\Im^{-1}$ denote the forward and inverse Discrete Fourier transforms, respectively. The outputs $O_{\mu,\nu}(z)$, where $\nu \in \{0,..., 4\}, \{\mu \in \{0,..., 7\}\}$, consist of different local, scale, and orientation features in both real and imaginary parts in the specific locality as described later in section 2.2. The magnitude of $O_{\mu,\nu}(z)$ is defined as modulus of $O_{\mu,\nu}(z)$, i.e. $\|O_{\mu,\nu}(z)\|$.

*2.2. Low-energized block based GWT feature extraction*

It is to be noted that we considered the magnitude of $O_{\mu,\nu}(z)$, but did not use the phase, which is consistent with the application of Gabor representations [14, 15]. As the outputs $(O_{\mu,\nu}(z): \mu\epsilon\{0,...,4\}, \nu\epsilon\{0,...,7\})$, consist of 40 different local scale and orientation features, the dimensionality of the Gabor transformed image space is very high. So the following technique is applied for the extraction of low energized discriminability feature vector $\chi_k$ from the convolution outputs. The method for the extraction of low-energized block based features from the GWT image is explained in Algorithm1.

Algorithm1:

Step 1: Find the convolution outputs of the original image with all the Gabor kernels. As the convolution outputs contain complex values, so replace each pixel value of the convolution output by its modulus and the resultant image is termed as $G_I$, where I=1,2,…,k, k= total Gabor kernels.

Step 2: Obtain the final single Gabor transformed image

$I_{GF} = \sum_{I=1}^{k} G_I, k = no.\,of\,Gabor\,kernels.$

Step 3: Compute the overall mean ($\bar{g}$) of the final Gabor transformed image $I_{GF}$ as, $\bar{g} = \frac{1}{m \times n}\sum_{x,y} I_{GF}(x,y)$, where $m \times n$ is the size of image.

Step 4: Divide the final Gabor transformed image $I_{GF}$ into windows of size $\omega \times \omega$. Thus the total number of windows, $l = \left\lfloor \frac{m}{\omega} \right\rfloor \times \left\lfloor \frac{n}{\omega} \right\rfloor$.

Step 5: For each window $w_i$, if minimum$(w_i) \leq \bar{g}$, then extract a block $B_i$ of size c×c from $w_i$, with centre pixel as the minimum $(w_i)$. The value of c must be odd integer and less than $\omega/2$.

Step 6: For each window $w_i$, if minimum$(w_i) \leq \bar{g}$, and there does not exists block $B_i$ of size c×c from $w_i$ as mentioned in step 5, with centre pixel as the minimum $(w_i)$, then create a block $B_i$ of size c×c, by considering the unavailable pixel values as $\bar{g}$.

Step 7: For each window $w_i$, if minimum$(w_i) > \bar{g}$, then create a pseudo block $B_i$ of size c×c with all elements as $\bar{g}$.

Step 8: Extract feature vector $f_i$ from each block $B_i$ in a systematic order, where $f_i$ contains all elements of the block $B_i$.

Step 9: Concatenate all the feature vectors $f_i$, $i=1,2,...,l$ to obtain the final feature vector $\chi$, which is the final extracted low- energized feature vector. This extracted feature vector $\chi$, encompasses the low valued discriminable elements of the Gabor transformed image, and the size of this feature vector is $[(Total\ no.of\ blocks) \times (size\ of\ the\ block)]$ which is much lower in dimension in comparison to the original image (dimension : $m \times n$), and the GWT image (dimension : $k \times m \times n$).

Thus this augmented Gabor feature vector encompasses most of the discriminable feature elements of the Gabor wavelet representation set, $S = (O_{\mu,\nu}(z): \mu \in \{0,...,4\}, \nu \in \{0,...,7\})$. The window size $\omega \times \omega$ is one of the important feature of the above algorithm, as it must be chosen small enough to capture most of the important features and large enough to avoid redundancy. Since it is observed that there are some windows each of whose minimum value is not less than the overall mean, so Step 7 is applied in order not to get stuck on a local minimum.

In the experiments we took a window and a block of size $7 \times 7$ and $3 \times 3$ respectively, to extract the low-energized feature vector. Thus the extracted facial features can be compared locally, instead of using a general structure, allowing us to make a decision from the parts of the face.

## 3. Generalized Kernel Discriminative Common Vector (KDCV) Method

Sometimes the discriminative common vectors are not distinct in the original sample space. In such cases one can map the original sample space to a higher-dimensional space $F$, where the new discriminative common vectors in the mapped space are distinct from one another. This is because a mapping, $\Phi: R^N \rightarrow F, x \rightarrow \phi(x)$, can map two vectors that are linearly dependent in the original sample space onto two vectors that are linearly independent in $F$. As the mapped space can have arbitrarily large, possibly infinite, dimensionality, hence it is reasonable to use the DCV Method.

Let $\Phi = \{\Phi(x_1^1), \Phi(x_2^1) ... \Phi(x_{N_1}^1), \Phi(x_1^2) ... \Phi(x_{N_c}^c)\}$

represent the matrix whose columns are the transformed training samples in $F$. Here c is the number of training classes; the i[th] class contains $N_i$ samples. The within-class scatter matrix $S_W^\Phi$, the between-class scatter matrix $S_B^\Phi$, and the total scatter matrix $S_T^\Phi$ in $F$ are given by:
$S_W^\Phi = (\phi - \phi G)(\phi - \phi G)^T$;
$S_B^\Phi = (\phi U - \phi L)(\phi U - \phi L)^T$;
$S_T^\Phi = (\phi - J_M \phi)(\phi - J_M \phi)^T = S_B^\Phi + S_W^\Phi$

where $\mu^\Phi$ is the mean of all samples, $\mu_i^\Phi$ is the mean of samples of the i[th] class in $F$. Here $G = diag(G_1, G_2, ..., G_c) \in R^{M \times M}$ is a block-diagonal matrix and each $G_i \in R^{N_i \times N_i}$ is a matrix with all its elements equal to $1/N_i$; $U = diag(u_1, u_2, ..., u_c)$ is a block-diagonal matrix and each $u_i \in R^{N_i \times 1}$ is a vector with all its elements equal to $1/N_i$; $L = diag(l_1, l_2, ..., l_c) \in R^{M \times C}$ is a block-diagonal matrix and each $l_i \in R^{M \times 1}$ is a vector with the entries $\sqrt{N_i}/M$; $J_M \in R^{M \times M}$ is a matrix with entries $1/\sqrt{M}$. The aim of the DCV algorithm is to acquire the optimal projection transform W in the null space of $S_W$ [16]:
$$J(W) = \arg\max_{W^T S_W W=0} |W^T S_B W| = \arg\max_{W^T S_W W=0} |W^T S_T W|. \quad (5)$$
The approach for computing this optimal projection vector is as follows:

Step1: Project the training set samples onto the range $R(S_T^\Phi)$ of $S_T^\Phi$ through the Kernel PCA.
Step 2: Find vectors $V$ that span the null space of $\tilde{S}_W^\Phi$.
Step 3: Remove the null space of $V^T \tilde{S}_B^\Phi V$ if it exists.
Step 4: Obtain the final projection matrix $W$, as $W = (\Phi - \Phi 1_M) U \Lambda^{-1/2} V L$, where $\Lambda$ is the diagonal matrix with non-zero eigenvalues, U, the associated matrix of normalized eigenvectors, and $V$ is the basis for the null space of $\tilde{S}_w^\phi$, here there are at most (C-1) projection vectors.

Let the common vector be $\Phi(x_{com}^i)$, then each of the feature vectors can be written as: $\Phi(x_m^i) = \Phi(x_{com}^i) + \Phi(x_{diff}^i)$, where $\Phi(x_{com}^i) \epsilon V^\perp$, and $\Phi(x_{m\_diff}^i) \epsilon V$.

Here $\Phi(x_{com}^i)$ and $\Phi(x_{m\_diff}^i)$ represent the common and different parts of $\Phi(x_m^i)$ separately. It has been proved by Gulmezoglu et al. [17] that for all samples of the i[th] class, their common vector parts are same. The common vector can be written as: $\Phi(x_{com}^i) = \Phi(x_m^i) - \Phi(x_{m\_diff}^i)$.
Thus, a set of common vectors is obtained as:
$Q = \{\Phi(x_{com}^1), \Phi(x_{com}^2) ... \Phi(x_{com}^c)\}$.
Compute the optimal projection transform. Let $\tilde{S}_{com}^\Phi$ denote the total scatter matrix of Q. $W_{com}$ is composed of the eigenvectors corresponding to the positive eigenvalues of $\tilde{S}_{com}^\Phi$. $W_{com}$ is designed to satisfy the criteria :
$J(W) = \arg\max_{W^T \tilde{S}_W W=0} W^T \tilde{S}_W W$
$J(W) = \arg\max_{W^T \tilde{S}_W W=0} W^T S_{com}^\Phi W$

$W_{diff}$ can be calculated from the different vectors. $W_{diff}$ is composed of the eigenvectors corresponding to the positive eigenvalues of $\tilde{S}_{T\_diff}^\Phi$. The optimal projection transform W is obtained as: $W = W_{diff} + W_{com}$ (6). Thus for each sample $\Phi(x)$ in the kernel space using the generalized nonlinear KDCV method, we construct W and then extract the kernel discriminative common and different vector $Y_{com}$ and $Y_{diff}$. Then, $Y_{com} = W_{com}^T \Phi(x)$, and $Y_{diff} = W_{diff}^T \Phi(x)$. Finally, $Y = Y_{diff} + Y_{com}$ (7).

Thus we obtain a new sample set Y corresponding to X. This sample set Y is used for image classification. All mathematical properties of the linear DCV carries over to the kernel DCV method with the modifications that are applied to the mapped samples, $\Phi(x_m^i)$, i=1,2,..., $c$ where $c$=1,2,..., $N_i$. After performing the feature extraction, all training set samples of each class typically give rise to a single distinct discriminative common vector.

*3.1. KDCV Approach using Cosine Kernel Function*

Let $\chi_1, \chi_2, ..., \chi_n \in \mathbb{R}^N$ be the data in the input space, and $\Phi$ be a nonlinear mapping between the input space and the feature space; $\Phi: R^N \longrightarrow F$. Generally three classes of kernel functions are used for nonlinear mapping: a) the polynomial kernels, b) the Radial Basis Function (RBF) kernels, c) the sigmoid kernels [18].
The RBF kernels, are also known as isotropic stationary kernels is defined by $\Phi : [0, \infty) \longrightarrow \mathbb{R}$ such that $k(x, x^{'}) = \Phi(\|x - x^{'}\|)$, where $x, x^{'} \in \chi \subset \mathbb{R}$, and $\|.\|$ is the norm operator. Normally a Gaussian function is preferred as the RBF, in most of the RBF kernels in pattern classification applications. The Gaussian function for RBF kernels is given by: $k(x, x^{'}) = exp\left(-\frac{\|x-x^{'}\|}{\sigma^2}\right)$. But the globally used RBF kernels yields dense Gram matrices, which can be highly ill-conditioned for large datasets.
So in this work the cosine kernel function is considered as the kernel function $\Phi$, defined by:
$\Phi(x,y) = \frac{\pi}{4} cos\left(\frac{\pi(x.y)}{2}\right)$, (8). This result can be expressed in terms of the angle $\theta$ between the inputs:
$\theta = cos^{-1}\left(\frac{\Phi(x,y)}{\sqrt{(\Phi(x,x))\Phi(y,y)}}\right)$. This shows that this kernel has a dependence on the angle between the inputs.
As a practical matter, we note that cosine kernels do not have any continuous tuning parameters (such as kernel width in RBF kernels), which can be laborious to set by cross validation.
Large margin classifiers are known to be sensitive to the way features are scaled [19]. Therefore it is essential to normalize either the data or the kernel itself. The recognition accuracy can severely degrade if the data is not normalized [19]. Normalization can be performed at the level of the input features or at the level of the kernel. It is often beneficial to scale all features to a common range, e.g., by standardizing the data. An alternative way to normalize is to convert each feature vector in to a unit vector. If the data is explicitly represented as vectors one can normalize the data by dividing each vector by its norm such that $\|x\| = 1$, after normalization. Here normalization is performed at the level of the kernel, i.e. normalizing in feature-space, leading to $\phi(x) = 1$ (or equivalently that $k(x, x) = 1$). This is accomplished by using the cosine kernel which normalizes a kernel $k(x, x^{'})$ to: $k_{cosine}(x, x^{'}) = \frac{k(x,x^{'})}{\sqrt{k(x,x).k(x^{'},x^{'})}}$. Normalizing data to unit vectors reduces the dimensionality of the data by one since the data is projected to the unit sphere.
In order to derive positive kernel nonlinear discriminating features (7), we consider only those eigenvectors that are associated with positive eigenvalues.

## 4. Similarity Measures and Classification

Finally the lower-dimensional, low energized extracted feature vector of the GWT image is used as the input data instead of the whole image in the proposed method to derive the kernel discriminative feature vector, $W$, using (6). Let $M_k^{'}$ be the mean of the training samples for class $w_k$, where $k = 1,2,..., l$ where $l$ is the number of classes. The classifier then applies, the nearest neighbor (to the mean) rule for classification using the similarity (distance) measure $\delta$ :

$\delta(Y, M_j^{'}) = \min_k (\delta(Y, M_k^{'})) \Rightarrow Y \in w_j$ (9)

The low energized KDCV vector $Y_L$ is classified to that class of the closest mean $M_k^{'}$ using the similarity measure $\delta$. The similarity measures used here are $L_1$ distance measure, $\delta_{L_1}$, $L_2$ distance measure, $\delta_{L_2}$, and the cosine similarity measure, $\delta_{cos}$, which are defined as:

$$\delta_{L_1} = \sum_i |X_i - Y_i| \quad (10)$$
$$\delta_{L_2} = (X-Y)^T(X-Y) \quad (11)$$
$$\delta_{cos} = \frac{-X^T Y}{\|X\|\|Y\|} \quad (12)$$

where T is the transpose operator and $\|.\|$ denotes the norm operator. Note that the cosine similarity measure includes a minus sign in (12) because the nearest neighbour (to the mean) rule (9) applies minimum (distance) measure rather than maximum.

*4.1. Experiments of the proposed method on frontal and pose-angled images for face recognition*

This section assesses the performance of the low-energized Gabor-block based KDCV method for both frontal and pose-angled face recognition. The effectiveness of the low-energized block based KDCV method is successfully tested on FRAV2D and FERET databases. For frontal face recognition, the data set is taken from the FRAV2D database, which consists of 1100 frontal face images corresponding to 100 individuals. The images are acquired, with partially occluded face features and different facial expression. For pose-angled face recognition, the data set taken from the FERET database contains 2200 images of 200 individuals with different facial expressions and poses. Further studies has been made on the FERET dataset using the standard protocols, i.e., the Fa, Fb, DupI, DupII set to assess the performance of the proposed method.

*4.1.1 FRAV2D Face Database:*

The FRAV2D face database, employed in the experiment consists; 1100 colour face images of 100 individuals, 11 images of each individual are taken, including frontal views of faces with different facial expressions, under different lighting conditions. All colour images are transformed into gray images and are scaled to a size of $m \times n$ here (92 ×

112) is used. The details of the images are as follows: (A) regular facial status; (B) and (C) are images with a 15° turn with respect to the camera axis; (D) and (E) are images with a 30° turn with respect to the camera axis. Images (F) and (G) are with gestures, such as smiles, open mouth, winks, laughs; (H) and (I) are images with occluded face features; (J) and (K) are images with change of illumination. Fig. 3 shows all samples of one individual.

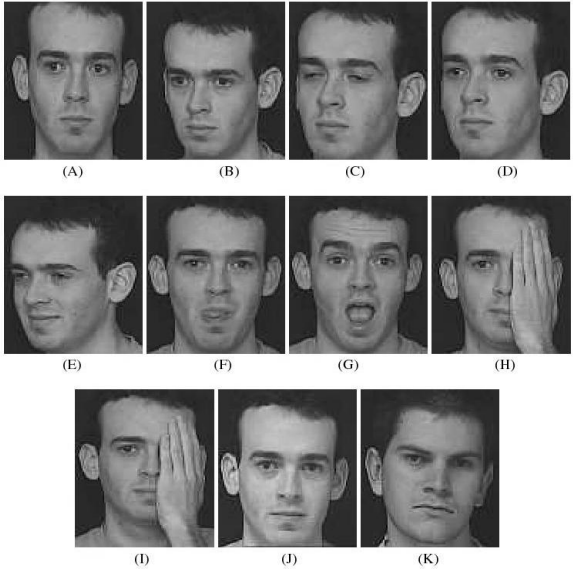

Fig. 1.Demonstration images of an individual from the FRAV2D database.

Table 1. Average recognition results using FRAV2D database:

| Method | Recognition Accuracy (%) No. of Training samples | | Average Recognition Rates(%) |
|---|---|---|---|
| | 3 | 4 | |
| GWT | 85.5 | 89.5 | 87.5 |
| KDCV | 79.5 | 82 | 80.75 |
| GWT-LDA | 88.3 | 90.33 | 89.33 |
| GWT- KDCV (RBF) | 87 | 90 | 88.5 |
| GWT-KDCV (COSINE) | 88 | 90 | 89 |
| GWT-KDCV (RBF, LOW- ENERGIZED) | 90 | 92.5 | 91.25 |
| GWT-KDCV (COSINE , LOW- ENERGIZED) | 96.125 | 97.75 | 96.9 |

*4.1.2. Specificity and Sensitivity measures for the FRAV2D dataset:*

To measure the sensitivity and specificity [20] the dataset from the FRAV2D database is prepared in the following manner. For each individual a single class is constituted with 18 images. Thus a total of 100 classes are obtained, from the dataset of 1100 images of 100 individuals. Out of the 18 images in each class, 11 images are of a particular individual, and 7 images are of other individuals taken by permutation as shown in Fig. 2. Using this dataset the true positive ($T_P$); false positive ($F_P$); true negative ($T_N$); false negative ($F_N$); are measured. From the 11 images of the particular individual, at first the first 4 images (A-D), then the first 3 images (A-C) of a particular individual are selected as training samples and the remaining images of the particular individual are used as positive testing samples. The negative testing is done using the images of the other individuals. Fig. 2 shows all sample images of one class of the data set used from FRAV2D database.

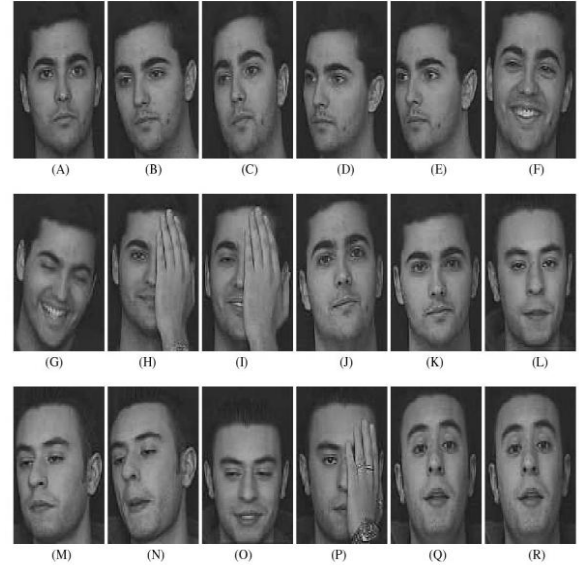

Fig. 2 Demonstration images of one class from the FRAV2D database

Table 2: Specificity and Sensitivity measure of the FRAV2D dataset:

| Total no. of classes=100, Total no. of images= 1800 | | | |
|---|---|---|---|
| | | Individual belonging to a particular class | |
| | | Using first 3 images of an individual as training images | |
| | | Positive | Negative |
| FRAV2D test | Positive | $T_P$ = 759 | $F_P$ =7 |
| | Negative | $F_N$ = 41 | $T_N$ =693 |
| | | Sensitivity = $T_P$ / ($T_P$ + $F_N$) ≈ 94.875% | Specificity =$T_N$/ ($F_P$ + $T_N$) =99.0% |
| | | Using first 4 images of an individual as training images | |
| | | Positive | Negative |
| FRAV2D test | Positive | $T_P$ = 678 | $F_P$ =2 |
| | Negative | $F_N$ = 22 | $T_N$ =698 |
| | | Sensitivity = $T_P$ / ($T_P$ + $F_N$) ≈ 96.85% | Specificity =$T_N$/ ($F_P$ + $T_N$) ≈99.7% |

So considering the first 4 images (A-D) of a particular individual for training the achieved rates are :
False positive rate = FP / (FP + TN) = 1 − Specificity =.3%
False negative rate = FN / (TP + FN) = 1 − Sensitivity=3.15%
**Accuracy = ($T_P$+$T_N$)/($T_P$+$T_N$+$F_P$+$F_N$) ≈ 98.3%.**
So considering the first 3 images (A-C) of a particular individual for training the achieved rates are:
False positive rate = FP / (FP + TN) = 1 − Specificity =1%
False negative rate = FN / (TP + FN) = 1 − Sensitivity=5.125%
**Accuracy = ($T_P$+$T_N$)/($T_P$+$T_N$+$F_P$+$F_N$) ≈ 96.9%.**

*4.1.3 FERET face database:*

The FERET database, employed in the experiment here contains, 2,200 facial images corresponding to 200 individuals with each individual contributing 11 images. The images in this database were captured under various illuminations, which display a variety of facial expressions and poses. As the images include the background and the body chest region, so each image is cropped to exclude those, and are transformed into gray images and is scaled to $m \times n$ here ($92 \times 112$) is used. Fig. 3 shows all samples of one subject. The details of the images are as follows: (A) regular facial status; (B) $+15°$ pose angle; (C) $-15°$ pose angle; (D) $+25°$ pose angle; (E) $-25°$ pose angle; (F) $+40°$ pose angle; (G) $-40°$ pose angle; (H) $+60°$ pose angle; (I) $-60°$ pose angle; (J) alternative expression; (K) different illumination.

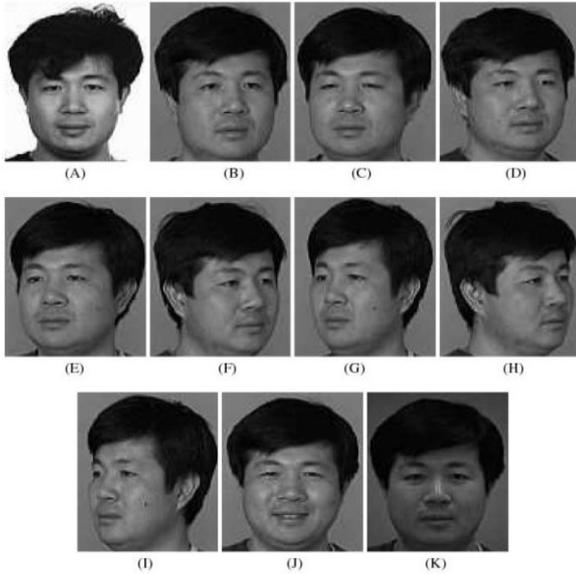

Fig. 3. Demonstration images of one individual from the FERET database.

Firstly 4 images of each individual i.e. A-D are regarded as training samples. The remainders are regarded as testing samples. After that 3 images of each individual i.e. A-C are regarded as training samples.

Table 3. Average recognition results using FERET database:

| Method | Recognition Rates (%) No. of Training samples | | Average Recognition Rates (%) |
|---|---|---|---|
| | 3 | 4 | |
| GWT | 79.65 | 82.5 | 81.1 |
| KDCV | 69.5 | 75 | 72.25 |
| GWT-LDA | 82.76 | 85.33 | 84.1 |
| **GWT- KDCV (RBF)** | 81 | 82.50 | 81.75 |
| **GWT-KDCV (COSINE)** | 83 | 85 | 84.5 |
| **GWT-KDCV (RBF, LOW- ENERGIZED)** | 88.5 | 91.5 | 90 |
| **GWT-KDCV (COSINE , LOW- ENERGIZED)** | 92.5 | 95.75 | 94.1 |

*4.1.4 Specificity and Sensitivity measure for the FERET dataset:*

To measure the sensitivity and specificity, the dataset from the FERET database is prepared in the following manner. For each individual a single class is constituted with 18 images. Thus a total of 200 classes are obtained, from the dataset of 2200 images of 200 individuals. Out of these 18 images in each class, 11 images are of a particular individual, and 7 images are of other individuals taken using permutation as shown in Fig. 4. Similarly using this dataset from the FERET dataset the specificity and sensitivity are being measured. From the 11 images of the particular individual, at first the first 4 images (A-D), then the first 3 images (A-C) of a particular individual are selected as training samples and the remaining images of the particular individual are used as positive testing samples. The negative testing is done using the images of the other individuals. Fig. 4 shows all sample images of one class of the data set used from FERET database.

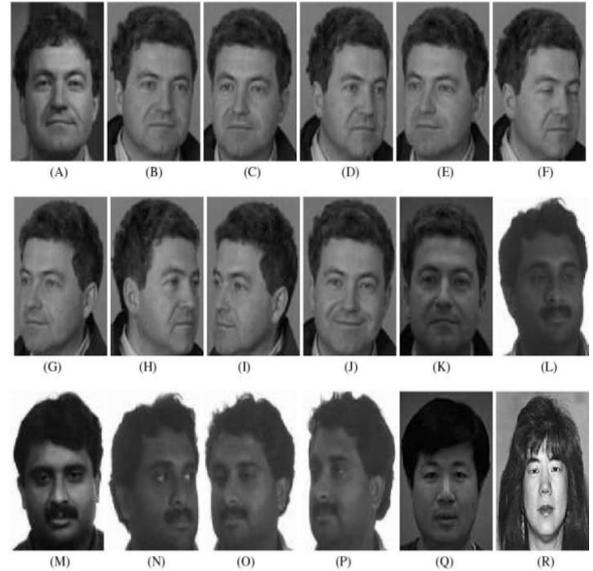

Fig. 4. Demonstration images of one class from the FERET dataset

Table 4: Specificity and Sensitivity measure of the FERET dataset:

| Total no. of classes=200, Total no. of images= 3600 | | | |
|---|---|---|---|
| | | Individual belonging to a particular class | |
| | | Using first 3 images of an individual as training images | |
| | | Positive | Negative |
| **FERET test** | Positive | $T_P = 1492$ | $F_P = 16$ |
| | Negative | $F_N = 108$ | $T_N = 1384$ |
| | | **Sensitivity = $T_P / (T_P + F_N) \approx 93.25\%$** | **Specificity = $T_N / (F_P + T_N) \approx 98.8\%$** |
| | | Using first 4 images of an individual as training images | |
| | | Positive | Negative |
| **FERET test** | Positive | $T_P = 1316$ | $F_P = 10$ |
| | Negative | $F_N = 84$ | $T_N = 1390$ |
| | | **Sensitivity = $T_P / (T_P + F_N) \approx 94\%$** | **Specificity = $T_N / (F_P + T_N) \approx 99.28\%$** |

So considering the first 4 images (A-D) of a particular individual for training the achieved rates are:
False positive rate = FP / (FP + TN) = 1 − Specificity = .72%
False negative rate = FN / (TP + FN) = 1 − Sensitivity= 6%
**Accuracy = $(T_P+T_N)/(T_P+T_N+F_P+F_N) \approx 96.6\%$.**
So considering the first 3 images (A-C) of a particular individual for training the achieved rates are:
False positive rate = FP / (FP + TN) = 1 − Specificity =1.2%
False negative rate = FN / (TP + FN) = 1 − Sensitivity= 6.75%
**Accuracy = $(T_P+T_N)/(T_P+T_N+F_P+F_N) \approx 96.1\%$.**

## 4.2 Further evaluation on the FERET Face dataset:

We further use the FERET face database for testing our proposed method, as it is one of the most widely used databases to evaluate face recognition algorithms [21]. FERET contains a gallery set, Fa, and four testing sets: Fb, Fc, DupI and Dup II. In the Fa set, there are 1196 images and contains one image per individual. The Fb set has 1195 images of people with different gestures. In the Fc set, there are 194 images under different lighting conditions. The DupI set has 722 images of pictures taken between 0 and 34 months of difference with those taken for Fa. The DupII set contains 234 images acquired at least 18 months after the Fa set. DupII is a subset of DupI. Fig. 4 shows samples from FERET face database. All images are aligned and cropped to $112 \times 92$ according to [22].

The extracted low-energized Gabor feature vector is considered as input to a trained KDCV with cosine kernel and its output is compared with a gallery set using the $L_1$, $L_2$ and cosine similarity measure. The recognition rates of different methods on the FERET probe sets are show in Table 5. The results are compared with the most recent state-of-the-art with the FERET database. Our results with the FERET database are equivalent (with a difference of $\pm 1\%$) to the most recent works on the FERET dataset. Note that the methods described in [23, 24] use the Gabor wavelets to generate their feature vectors. As the Gabor wavelets have a much higher algorithmic complexity, so the overall computing cost is very high. On the other hand, our block-based low energized Gabor features is a very low-dimensional feature vector which reduces the algorithmic complexity. Also with the use of cosine kernel as the kernel function in the KDCV makes the proposed method quite fast and more suitable to real applications.

Table 5. Recognition results of different algorithms on the FERET Probe Sets:

| Method | FERET Probe Sets | | | |
|---|---|---|---|---|
| | Fb (%) | Fc (%) | Dup I (%) | Dup II (%) |
| Phillips et. al. | 96 | 82 | 59 | 52 |
| Local Gabor binary pattern histogram sequence, [23] | 98 | 97 | 74 | 71 |
| Grassmann registration manifolds for face recognition, [24] | 98 | 98 | 80 | 84 |
| Low-energized Gabor Block based KDCV with RBF Kernels (Gauss) using Cosine measure. | 96 | 97 | 79 | 81 |
| Low-energized Gabor Block based KDCV with cosine Kernels using $L_1$ measure. | 95 | 96 | 79 | 80 |
| Low-energized Gabor Block based KDCV with cosine Kernels using $L_2$ measure. | 96 | 97 | 86 | 82 |
| **Low-energized Gabor Block based KDCV with cosine Kernels using Cosine measure.** | **98** | **98** | **89** | **84** |

Furthermore, we compare the face recognition performance of our proposed low- energized GWT-KDCV method using cosine kernels, with some other well known methods like generalized discriminant analysis (GDA) method [7], (EGM) [9], Discrete Cosine transformation (DCT) and linear disrminant analysis (LDA), DCT-LDA method [25], DCT-GDA [20], GWT-LDA, DCT-KDCV method [26], and the Gabor fusion KDCV method [27]. Classification results obtained of the proposed method are comparable or even better in some cases than above mentioned methods. Also the cosine similarity measure is more suitable for classifying the non linear real KDCV features shown in the tables, table 1 to table 6.

Table 6. Comparison of Recognition accuracy of various methods with the proposed method.

| Method | Highest Recognition Accuracy |
|---|---|
| GDA | 78.04% |
| Elastic graph matching(EGM) [9] | 80.00% |
| DCT-LDA | 80.87% |
| GWT | 81.1% |
| DCT-GDA | 82.84%, |
| GWT-LDA | 84.1% |
| DCT-KDCV | 85.13% |
| Gabor fusion KDCV | 91.22 |
| **Proposed Approach** | **94.25%** |

As the proposed method performs best with the cosine similarity classifier, so the specificity rate of the proposed method is evaluated for the FERET and FRAV2D dataset using the cosine similarity measure shown in Fig. 7.

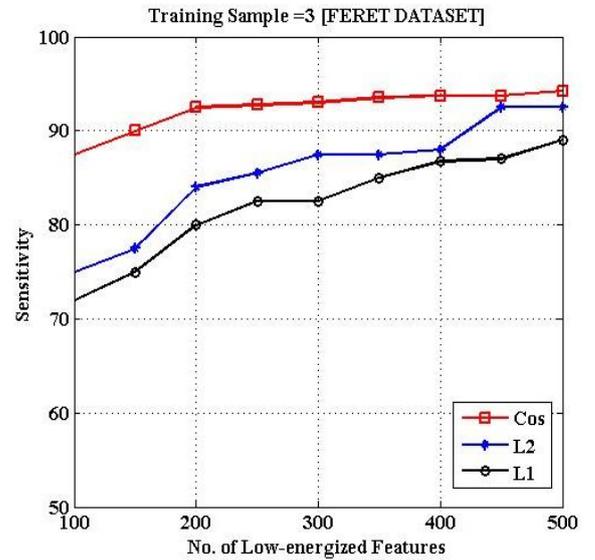

Fig. 5. Face recognition performance of the proposed method using cosine kernel function, and considering the first 3 images as training set on the FERET dataset, with the three different similarity measures: cos ( cosine similarity measure), $L_2$ ($L_2$ distance measure), $L_1$ ( $L_1$ distance measure).

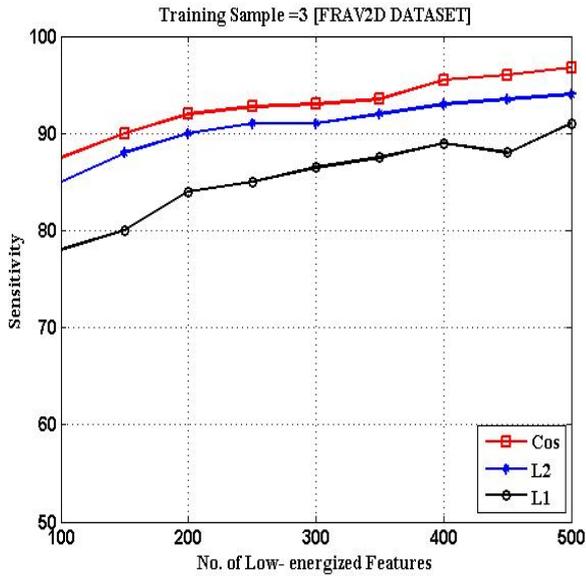

Fig. 6. Face recognition performance of the proposed method using cosine kernel function, and considering the first 3 images as training set on the FRAVD2D dataset, with the three different similarity measures: cos ( cosine similarity measure), $L_2$ ($L_2$ distance measure), $L_1$ ( $L_1$ distance measure).

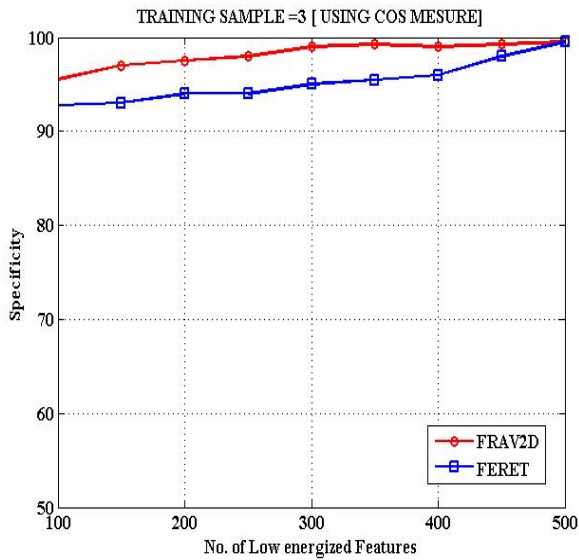

Fig. 7. Negative recognition performance of the proposed method using Cosine similarity distance measure on the FRAV2D and FERET dataset.

## 4.3 Results

Experiments conducted using the low-energized block based Gabor KDCV method, with three different similarity measures on both the FERET and FRAV2D databases are shown in Fig. 5 and Fig. 6. Considering, only the low dimensional, low energized features of GWT image, greatly improves the computing speed of nonlinear discriminant method. The results of recognition accuracy (in terms of sensitivity) versus dimensionality reduction (number of features) and the cumulative match curves using the three different similarity measures are shown in Fig. 5 and Fig. 6. From the results on both FERET and FRAV2D database it is seen that the cosine similarity measure performs the best, followed in order by the L2 and the L1 measure.

Fig. 5 and 6 indicates that the proposed method performs well with a lower dimension as well. These results show that there is certain robustness to age and illumination. Our results indicate that:

i) the low-energized block based Gabor features with KDCV approach greatly enhance the face recognition performance as well as reduce the dimensionality of the feature space when compared with the Gabor features as shown in Table 1 and 3. For example, the similarity measure improves the face recognition accuracy by almost 10% using only the few low-energized Gabor features with improved discriminative power compared to the original Gabor features as shown in Table 1and 3.

ii) The proposed method further enhances face recognition with the use of cosine kernel along with the cosine similarity measure.

Experimental result indicates that the use of cosine kernels in the KDCV further increases the discriminative power of the feature vector extracted from the low- energized block of GWT image and hence, is an effective feature extraction approach, performing better way to extract more effective discriminating features than the GDA. The extracted low-energized feature vector by the proposed Algorithm1 of section 2.2 enhances the face recognition performance in presence of occlusions. Experimentally it has been observed that this method is less time consuming than the EGM and other well known algorithms [22, 23, 24, 25, 26]. a) Our results show that the proposed method greatly enhances recognition performance, b) reduces the dimensionality of the feature space, c) the Cosine similarity distance classifier further increases the face recognition accuracy, as it calculates the angle between two vectors and is not affected by their magnitude.

## 5 Conclusion

This paper introduces a novel block-based GWT generalized KDCV method using the cosine kernel for frontal and pose-angled face recognition. As cosine kernel function is used here, so there is no need of data normalization and the parameter tuning can be avoided as in the case of (RBF) Gaussian kernels. Also the derived low dimensional low-energized features are characterized by spatial frequency, locality, and orientation selectivity to cope with the variations due to illumination and facial expression changes as a property of Gabor kernels. Such characteristics produce salient local features, such as the eyes, nose and mouth, that are suitable for face recognition. The KDCV method extended with cosine kernel is then applied on these extracted feature vectors to finally obtain only the real nonlinear discriminating kernel feature vector with improved discriminative power, containing salient facial features that are compared locally instead of a general structure, and hence allows to make a decision from the different parts of a face, and thus maximizes the benefit of applying the idea of "recognition by parts". So the method performs well in presence of occlusions (e.g., sunglass, scarf etc.) that is when there are sunglasses or any other obstacles the algorithm compares face in terms of mouth, nose and other features rather than the eyes.

**Acknowledgement** Authors are thankful to a major project entitled "Design and Development of Facial Thermogram Technology for Biometric Security System," funded by University Grants Commission (UGC),India and "DST-PURSE Programme" and CMATER and SRUVM project at Department of Computer Science and Engineering, Jadavpur University, India for providing necessary infrastructure to conduct experiments relating to this work.